\documentclass[sigconf]{acmart}

\AtBeginDocument{%
  }

\setcopyright{acmcopyright}
\copyrightyear{2025}
\acmYear{2025}
\setcopyright{cc}
\setcctype{by}
\acmConference[KDD '25]{Proceedings of the 31st ACM SIGKDD Conference on
Knowledge Discovery and Data Mining V.2}{August 3--7, 2025}{Toronto, ON,
Canada}
\acmBooktitle{Proceedings of the 31st ACM SIGKDD Conference on Knowledge
Discovery and Data Mining V.2 (KDD '25), August 3--7, 2025, Toronto, ON,
Canada}
\acmDOI{10.1145/3711896.3736560}
\acmISBN{979-8-4007-1454-2/2025/08}

\newcommand{\Escr}{\ensuremath{\mathcal E}}
\newcommand{\Fscr}{\ensuremath{\mathcal F}}

\newcommand{\Mscr}{\ensuremath{\mathcal M}}

\def\P{\mathbb{P}}
\def\Q{\mathbb{Q}}

\usepackage{stackengine}

\begin{document}

\title{Data Heterogeneity Modeling for Trustworthy Machine Learning}


\author{Jiashuo Liu}
\email{liujiashuo77@gmail.com}
\affiliation{%
  \institution{Tsinghua University}
  \city{Beijing}
  \country{China}
}

\author{Peng Cui}
\email{cuip@tsinghua.edu.cn}
\affiliation{%
  \institution{Tsinghua University}
  \city{Beijing}
  \country{China}
}

\begin{abstract}
Data heterogeneity plays a pivotal role in determining the performance of machine learning (ML) systems. 
Traditional algorithms, which are typically designed to optimize average performance, often overlook the intrinsic diversity within datasets.
This oversight can lead to a myriad of issues, including unreliable decision-making, inadequate generalization across different domains, unfair outcomes, and false scientific inferences. 
Hence, a nuanced approach to modeling data heterogeneity is essential for the development of dependable, data-driven systems.
In this survey paper, we present a thorough exploration of heterogeneity-aware machine learning, a paradigm that systematically integrates considerations of data heterogeneity throughout the entire ML pipeline—from data collection and model training to model evaluation and deployment. 
By applying this approach to a variety of critical fields, including healthcare, agriculture, finance, and recommendation systems, we demonstrate the substantial benefits and potential of heterogeneity-aware ML. 
These applications underscore how a deeper understanding of data diversity can enhance model robustness, fairness, and reliability and help model diagnosis and improvements.
Moreover, we delve into future directions and provide research opportunities for the whole data mining community, aiming to promote the development of heterogeneity-aware ML. 
\end{abstract}



\maketitle

\section{Introduction}

\begin{figure*}[ht]
	\includegraphics[width=0.6\textwidth]{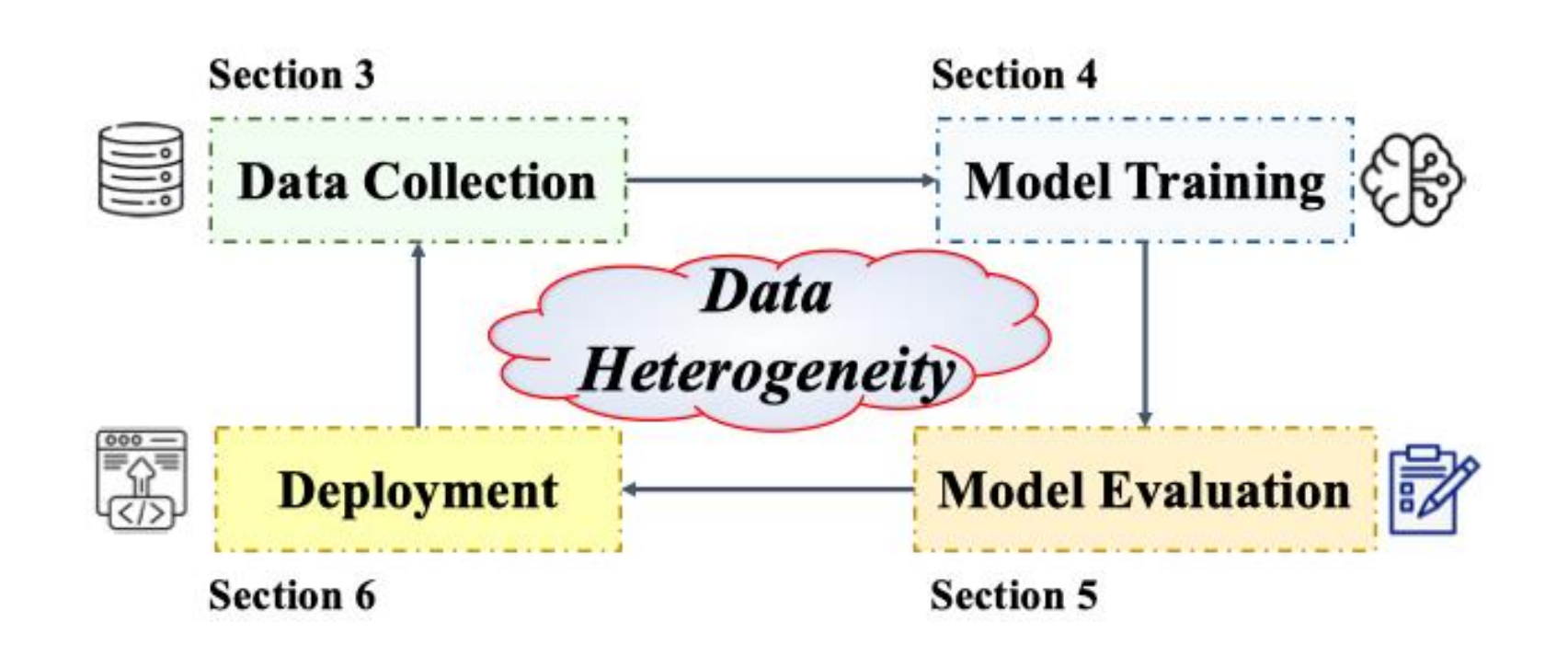}
    \vspace{-0.1in}
	\caption{Scope of heterogeneity-aware machine learning, which involves the whole machine learning pipeline and connects various high-stakes applications.}
	\label{fig:scope}
	\vspace{-0.1in}
\end{figure*}

Big Data provides great opportunities for the growth and advancement of Artificial Intelligence (AI) systems. 
Nowadays, AI has emerged as a ubiquitous tool that permeates almost every aspect of the contemporary technological landscape, making it an indispensable asset in various fields and industries, such as scientific discoveries, policy-making, healthcare, drug discovery, and so on.
However, along with the widespread deployment of AI systems, the reliability, fairness, and stability of AI algorithms have been increasingly doubted.
For example, in sociological research \cite{2020Toward}, studies have shown that even for carefully designed randomized trials, there are huge selection biases, making scientific discoveries unreliable; in disease diagnosis \cite{wynants2020prediction, roberts2021common}, studies have found hundreds of existing AI algorithms fail to detect and prognosticate for COVID-19 using chest radiographs and CT scans; in social welfare, decision support AI systems for credit loan applications are found to exhibit biases against certain demographic groups \cite{hardt2016equality, verma2019weapons}; in various machine learning tasks, algorithms are faced with severely poor generalization performances under distributional shifts \cite{shen2021towards}.

In order to mitigate the barriers against AI systems in high-stakes applications, numerous researchers have made efforts following the established research paradigm of model-centric AI, where they design innovative algorithms to enhance the generalization and reliability.
For example, distributionally robust optimization (DRO) methods \cite{duchi2021learning} propose to optimize the worst-case distribution lying around the training distribution to guarantee the performances under unexpected cases in testing;
and invariant learning methods \cite{arjovsky2019invariant} instead aim to learn invariant prediction mechanisms across heterogeneous environments. 
Despite the intellectual appeal and theoretical promise of these algorithms, their translation into practical benefits in real-world applications has been limited. 
Empirical studies on image data \cite{gulrajani2020search} and tabular data \cite{liu2024need} have illustrated such discrepancy between theoretical robustness and empirical effectiveness.
This gap largely stems from an oversight in thoroughly investigating the properties of the data used for developing ML models. 
More specifically, many of these methods presuppose certain data characteristics without rigorous validation, leading to a gap between their theoretical assumptions and practical utility.

In current machine learning, it is increasingly evident that the challenges faced by algorithms extend beyond their intrinsic properties and extend to the nature of the data utilized in training these models. 
Specifically, the heterogeneity of data employed has emerged as a pivotal factor underlying these issues.
The concept of data heterogeneity encompasses the \emph{diversity} that exists within data, including \emph{variations in data sources, different generating processes, latent sub-populations, etc} \cite{fan2014challenges}. 
Failure to account for such diversity in AI systems can lead to overemphasis on patterns found only in dominant sub-populations or groups, thereby resulting in false scientific discoveries, unreliable and inequitable decision-making, and poor generalization performance when confronted with data from the minority groups. 
For instance, historical data from the U.S. credit market reveals that minorities have faced systematic disparities, such as higher denial rates for loans, mortgages, and credit cards, or being subjected to higher interest rates compared to other consumers \cite{engel2008credit, WinNT}.
Likewise, a study \cite{obermeyer2019dissecting} found that an algorithm commonly employed in U.S. hospitals for distributing health care resources to patients has consistently shown bias against black individuals.
Therefore, for these high-stakes scenarios where trustworthy AI is required, addressing the problem of data heterogeneity - an inherent property of big data - should receive increased attention. 
Moreover, in the current era of big models, where model development is approaching its limit, \emph{data mining researchers have a unique opportunity to explore the intricacies of big data}, thereby facilitating the development of AI in parallel with the advancement of AI models and algorithms.

Recently, data heterogeneity has attracted considerable attention across various disciplines, including statistics \cite{fan2014challenges}, medicine science \cite{dahabreh2016using, kent2018personalized}, causal inference \cite{athey2019generalized,wager2018estimation}, and machine learning \cite{liu2021heterogeneous, liu2023}, etc.
Although these studies share similar principles, there is a lack of a unified approach to studying it in machine learning.
This study represents a pioneering effort to offer a comprehensive and integrated perspective on \textbf{H}eterogeneity-\textbf{A}ware \textbf{M}achine \textbf{L}earning (HAML). 
Our survey aims to systematically incorporate data heterogeneity throughout the entire machine learning pipeline, encompassing data collection, model training, evaluation, and deployment phases, as shown in Figure \ref{fig:scope}.
We will demonstrate how the principle of data heterogeneity can be seamlessly integrated at different stages of the ML pipeline. 
Furthermore, we highlight the significant advantages that such an approach can offer, particularly in the context of real-world, high-stakes applications, thereby underlining the critical need for a unified treatment of data heterogeneity in machine learning.
The main body of this survey is structured as follows:
\begin{enumerate}
	\item Section \ref{sec:pre} (Preliminaries): Provides a critical review of traditional ``model-centric'' methodologies in machine learning, advocating for the transition to a heterogeneity-aware approach to ensure ML trustworthiness.
	\item Section \ref{sec:collection} (Data Collection): Introduces and defines the concept of predictive heterogeneity, demonstrating its utility in exploring and modeling dataset structures, with specific examples from healthcare and agriculture.
	\item Section \ref{sec:training} (Model Training): Discusses the integration of data heterogeneity into the model training process, highlighting its critical role, particularly in the applications of graph data and recommendation systems.
	\item Section \ref{sec:evaluation} (Model Evaluation): Explores the advantages of considering data heterogeneity during model evaluation, including different evaluation metrics and algorithms, as well as application in healthcare and face recognition.
	\item Section \ref{sec:deployment} (Model Deployment): Discusses the impact of a better understanding of data heterogeneity in addressing and rectifying model failures after deployment, complemented by an in-depth healthcare case study.
\end{enumerate}

Throughout these sections, we incorporate instances of high-stakes real-world applications to illustrate the extensive potential and advantages of embracing a heterogeneity-aware machine learning paradigm.

\section{Preliminaries}
\label{sec:pre}

\textbf{Notations}.\quad Throughout this paper, we let $\mathbb R$ denote the set of real numbers, $\mathbb R_+$ denote the subset of non-negative real numbers.
We use capitalized letters for random variables, e.g., $X,Y,Z$, and script letters for the sets, e.g., $\mathcal X, \mathcal Y, \mathcal Z$.
All random variables are defined in a probability space $(\Omega, \mathcal F, \mathbb P)$.
For any close set $\mathcal Z \subset \mathbb R^d$, we define $\mathcal P(\mathcal Z)$ as the family of all Borel probability measures on $\mathcal Z$.
For $\P \in \mathcal P(\mathcal Z)$, we use the notation $\mathbb E_\P[\cdot]$ to denote expectation with respect to the probability distribution $\P$.
For the prediction problem, the random variable of data points is denoted by $Z=(X,Y)\in \mathcal Z$, where $X\in \mathcal X$ denotes the input covariates, $Y\in\mathcal Y$ denotes the target. 
$f_\beta:\mathcal X\rightarrow \mathcal Y$ denotes the prediction model parameterized by $\beta$.
The loss function is denoted as $\ell:\mathcal Y \times \mathcal Y \rightarrow \mathbb R_+$, and $\ell(f_\beta(X),Y)$ is abbreviated as $\ell(\beta,Z)$. 
$E\in \Escr$ denotes an environment, and the data distribution in environment $E$ is denoted as $P_E(Z)$.

Great efforts have been made to mitigate the bias and generalization problems of ML systems.
Among diverse algorithms developed \cite{shen2021towards}, distributionally robust optimization (DRO) and invariant learning stand out as two prominent approaches.
In this section, we aim to provide a succinct overview of these methodologies, and to analyze the reasons why their effectiveness is limited in real-world scenarios.
For a thorough review of this line of research, we refer the readers to these survey papers \cite{shen2021towards,blanchet2024distributionally}.

DRO methods take the form of:
\begin{equation}
\label{equ:dro}
	\min_\beta\sup_{\Q:\mathcal{M}(\Q,\P_{\text{tr}})\leq \epsilon} \mathbb{E}_\Q[\ell(\beta,Z)],
\end{equation}
where $\P_{\text{tr}}$ denotes the training distribution, $\Mscr(\cdot,\cdot)$ denotes some distance metrics or divergences between distributions, like Wasserstein distance \cite{mohajerin2018data, blanchet2019robust}, $f$-divergence \cite{duchi2021learning}, MMD distance \cite{staib2019distributionally}, etc, and $\epsilon>0$ denotes the radius of the ambiguity set.
The core idea of DRO is to optimize the worst-case distribution lying around $\P_{\text{tr}}$, in preparation for future distribution shifts.
To be more specific, if the target distribution falls into the ambiguity set, DRO methods could ``theoretically'' guarantee the generalization performances.
Based on the formulation in Equation \eqref{equ:dro}, various tractable optimization methods are developed~\cite{liu2025dropythonlibrary}, and the relationships between DRO and regularizations \cite{mohajerin2018data, blanchet2019robust}, tilted empirical risk functions \cite{li2023tilted}, variance penalties \cite{duchi2021learning} are established.

Different from DRO methods which perturb the training distribution, invariant learning \cite{arjovsky2019invariant} assumes the invariant prediction structure across multiple data sources.
More specifically, the invariance assumption \cite{arjovsky2019invariant} is made that a representation $\Phi(X)$ is invariant if for any  $E_1, E_2\in\Escr, \mathbb E[Y|\Phi(X), E_1]=\mathbb E[Y|\Phi(X), E_2]$.	
And the goal is to learn such invariant representations via surrogated risk functions.
Follow-up works \cite{ahuja2020invariant, ahuja2021invariance, chen2022pareto} propose slightly different notions of invariance, and develop corresponding algorithms. 

\begin{figure}[t]
	\includegraphics[width=0.46\textwidth]{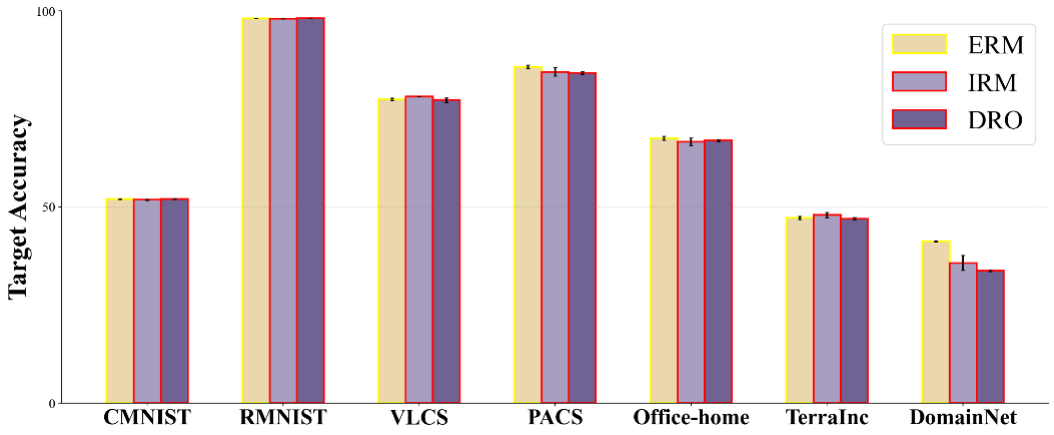}
    \vspace{-0.1in}
	\caption{Target accuracy on typical out-of-distribution generalization datasets for image data. Figure generated from \cite[Table 4]{gulrajani2020search}.}
	\label{fig:domainbed}
    \vspace{-0.2in}
\end{figure}

Although these algorithms hold theoretical appeals and promises, their practical application and benefits in real-world settings have been limited.
Empirical studies in DomainBed \cite{gulrajani2020search} illustrate that both Group DRO \cite{sagawa2019distributionally} and IRM \cite{arjovsky2019invariant} do not show improvements over empirical risk minimization, as shown in Figure \ref{fig:domainbed}.
Similar trends are also found on tabular data \cite{liu2024need}, where DRO methods do not outperform basic methods (e.g., Logistic Regression, SVM) and tree-based methods (e.g., random forest, XGBoost).

This discrepancy primarily arises from a lack of comprehensive analysis of the data characteristics upon which machine learning models are developed.
Specifically, many of these approaches make assumptions about data properties \emph{without} careful verification, resulting in a misalignment between their theoretical underpinnings and practical effectiveness.
For instance, as highlighted in \cite{liu2021heterogeneous}, when the predefined multiple environments are inaccurately specified, the resulting learned invariance property proves to be insufficient.
This underscores the importance of not merely making modeling assumptions and developing corresponding methods. 
Rather, it is essential to first comprehensively understand the application and its data. 
Only after gaining this understanding should we formulate suitable modeling assumptions or design algorithms. 
This paves the way for Heterogeneity-Aware Machine Learning.
\section{Data Collection}
\label{sec:collection}

The first stage in the machine learning (ML) pipeline is data collection, which encompasses both the gathering and pre-processing of data prior to model design and training.
To ensure the practical utility of ML algorithms, it is essential not to make assumptions blindly. Instead, we must first develop a thorough understanding of both the application context and the nature of the data itself.

In this section, we explore how data heterogeneity informs the following fundamental question:
\begin{center}
\bf Q1: How can we understand the data at hand?
\end{center}

\noindent Understanding data involves multiple dimensions—for example, identifying whether the dataset consists of distinct sub-populations, uncovering latent sub-structures, or detecting variations in noise levels across samples. A growing body of research has proposed methods to characterize such properties.
Here, we review approaches that specifically target the analysis of noise levels and data sub-populations.

\subsection{Noise Level Analysis}
The characterization of data quality through model training dynamics is a burgeoning field in machine learning. Methodologies such as Dataset Cartography \citep{Swayamdipta2020EMNLP} leverage metrics like confidence and its variability across epochs to map datasets into regions of easy-to-learn, hard-to-learn (often indicative of label errors), and ambiguous samples, the latter being crucial for out-of-distribution generalization. Similarly, techniques focusing on the Area Under the Margin (AUM) \citep{Pleiss2020NeurIPS} track logit margins to identify mislabeled data by observing conflicting signals during training. Further research explores other dynamic signals, including forgetting statistics \citep{Toneva2019ICLR}, which identify samples that models repeatedly learn and then misclassify, often correlating with noisy or atypical data. Influence functions \citep{Koh2017ICML} provide another avenue by estimating the impact of individual data points on model predictions or parameters, aiding in the identification of highly influential or detrimental samples \citep{Yoon2020ICML_DVRL, Grosse2023InfluenceExpensive}. 

Complementing these, frameworks like Data-IQ by \citet{Seedat2022NeurIPS} specifically address tabular data by analyzing training dynamics such as aleatoric uncertainty and predictive confidence to characterize subgroups (Easy, Ambiguous, Hard) with heterogeneous outcomes, thereby enhancing data understanding and model performance.
Related work also includes DIPS, which uses learning dynamics for selecting useful samples in pseudo-labeling \citep{Seedat2024DIPS}, and TRIAGE for characterizing training data in regression tasks \citep{Seedat2024TRIAGE}.
Collectively, these methods underscore a paradigm shift towards data-centric AI, where understanding and improving data quality through its interaction with the learning process is paramount for developing robust and efficient models \citep{Sambasivan2021DataCascades, DataRater2025Arxiv}.

While the above data-centric methods provide valuable insights into data quality, they lack a principled approach for uncovering latent data sub-populations. To address this gap, we introduce data sub-population analysis methods as follows.

\subsection{Data Sub-population Analysis}
Data sub-populations refer to latent groups within a dataset that may stem from different generative processes or exhibit distinct predictive behaviors. Identifying and understanding these sub-populations is crucial, especially in settings where model performance and fairness can be significantly impacted by data heterogeneity.

Several empirical methods have been proposed to uncover such sub-populations. For example, Liu et al.\cite{liu2021just} suggest treating misclassified samples as a distinct subgroup to improve model robustness. Creager et al.\cite{creager2021environment} take a different approach, leveraging the invariance penalty framework introduced by Arjovsky et al.~\cite{arjovsky2019invariant} to detect subgroups that violate assumed invariance across environments. 
While these methods offer practical insights and have shown empirical success, they generally lack a strong theoretical grounding to explicitly model the presence and structure of data sub-populations.

To address this gap, we highlight a seminal work that provides a principled approach to modeling data sub-populations. Specifically, Liu et al.\cite{liu2023} introduce the concept of predictive heterogeneity—formally defined in Definition\ref{def:ph}—to capture variations in predictive mechanisms across different subgroups within the data.

\begin{definition}[Predictive Heterogeneity \cite{liu2023}]
	Let $X$, $Y$ be random variables taking values in $\mathcal X \times \mathcal Y$ and $\mathscr E$ be a partition set, where each element is a discrete distribution $\Escr$ over the environment variable $E$. For a predictive family $\mathcal V$, the predictive heterogeneity for the prediction $X \rightarrow Y$ with respect to $\mathscr E$ is defined as:
	\begin{equation}
	\label{equ:usable-predictive-heterogeneity-1}
		\mathcal{H}^\mathscr E_{\mathcal V}(X \rightarrow Y) = \sup_{\Escr \in \mathscr E}\mathbb{I}_{\mathcal{V}}(X\rightarrow Y|\Escr)-\mathbb{I}_{\mathcal{V}}(X\rightarrow Y),
	\end{equation} 
	where $\mathbb{I}_{\mathcal{V}}(X\rightarrow Y|\Escr)$ is the conditional predictive $\mathcal V$-information defined as:
	\begin{align*}
		\mathbb{I}_{\mathcal{V}}(X\rightarrow Y|\mathcal{E}) &= H_{\mathcal V}(Y|\emptyset,\mathcal E)-H_{\mathcal V}(Y|X,\mathcal E),\\
		H_{\mathcal V}(Y|X,\Escr) &= \mathbb E_{E \sim \mathcal E} \left[ \inf\limits_{f\in\mathcal{V}}\mathbb{E}_{x,y\sim X,Y|\mathcal E=E}[-\log f[x](y)]\right], \\
		H_{\mathcal V}(Y|\emptyset,\mathcal E) &= \mathbb E_{E \sim \mathcal E} \left[ \inf\limits_{f\in\mathcal{V}}\mathbb{E}_{y\sim Y | \mathcal E=E}[-\log f[\emptyset](y)] \right].
	\end{align*}
	\label{def:ph}
\end{definition}

\begin{figure}[t]
 \centering\captionsetup[subfloat]{labelfont=scriptsize,textfont=scriptsize}
 \stackunder[3pt]{\includegraphics[width=0.49\linewidth]{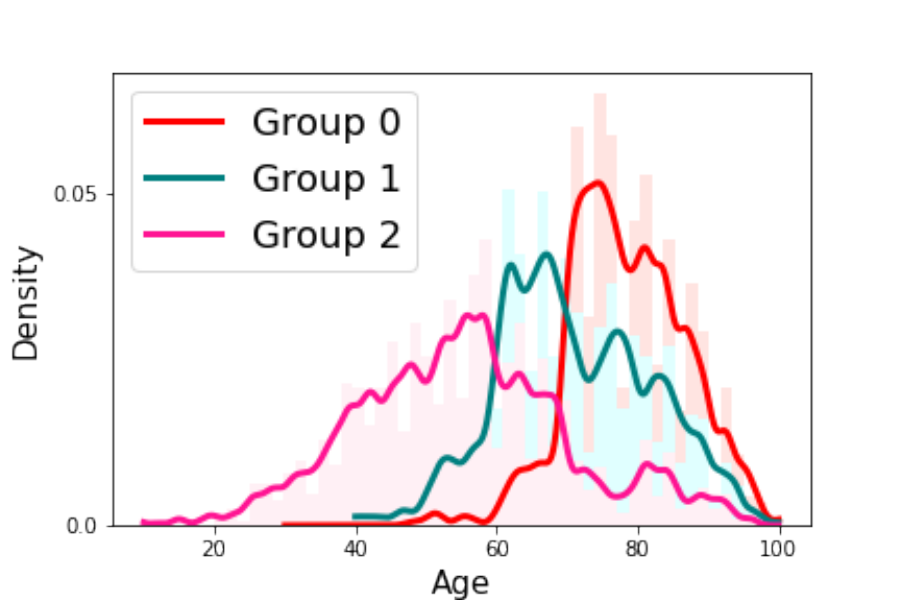}}{(a) Age distribution.}
 \stackunder[3pt]{\includegraphics[width=0.5\linewidth]{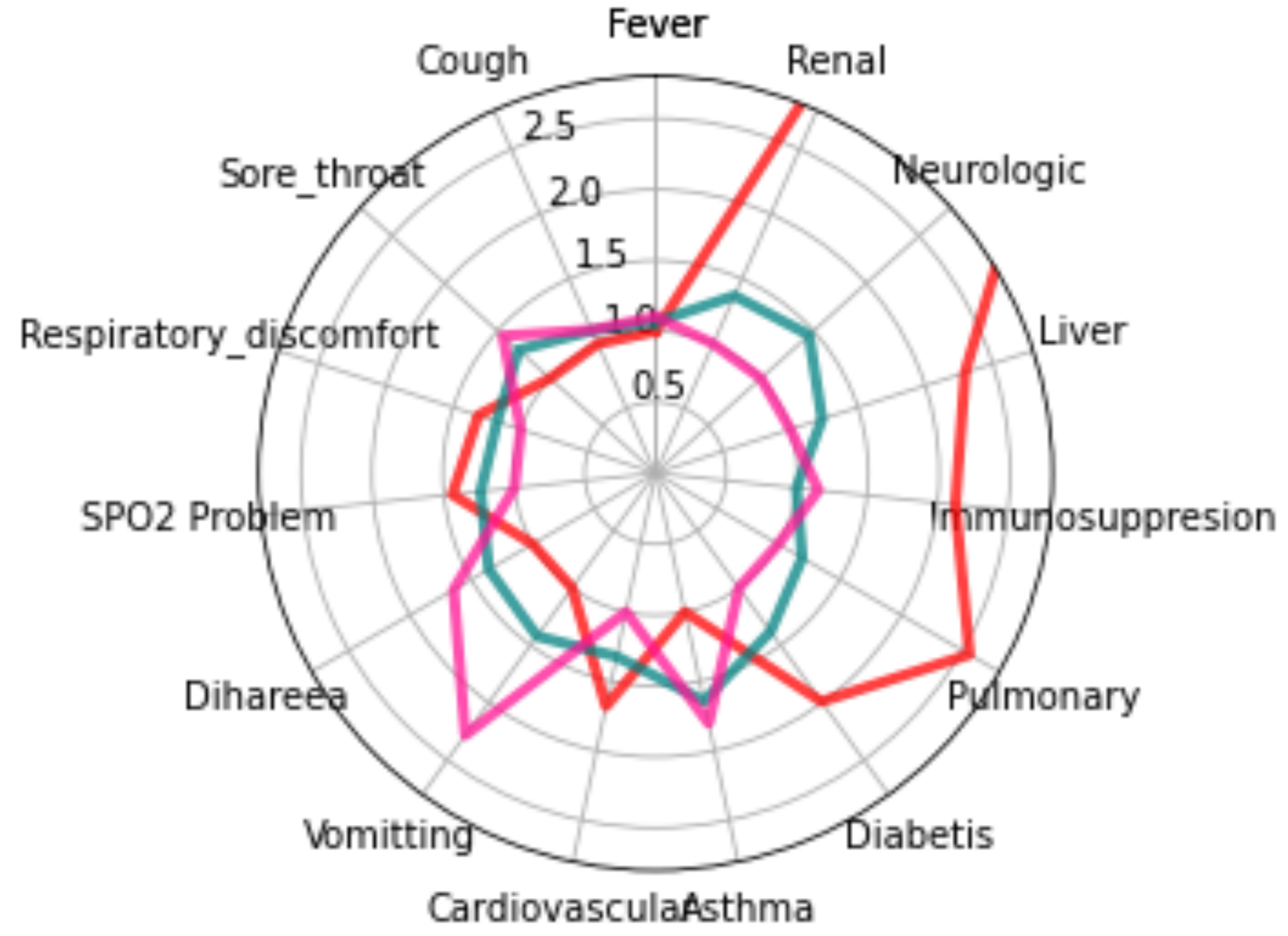}}{(b) Feature average.}
 \vspace{-0.1in}
\caption{Results on the COVID-19 data. (a): The age distributions of dead people ($Y=1$) in each learned subgroup. (b): The averages of typical features of dead people ($Y=1$) in each learned subgroup. Figures are from \cite{liu2023}.}
 \label{fig:covid}  
\vspace{-0.2in}
\end{figure}

\begin{figure*}[h]
	\includegraphics[width=0.75\textwidth]{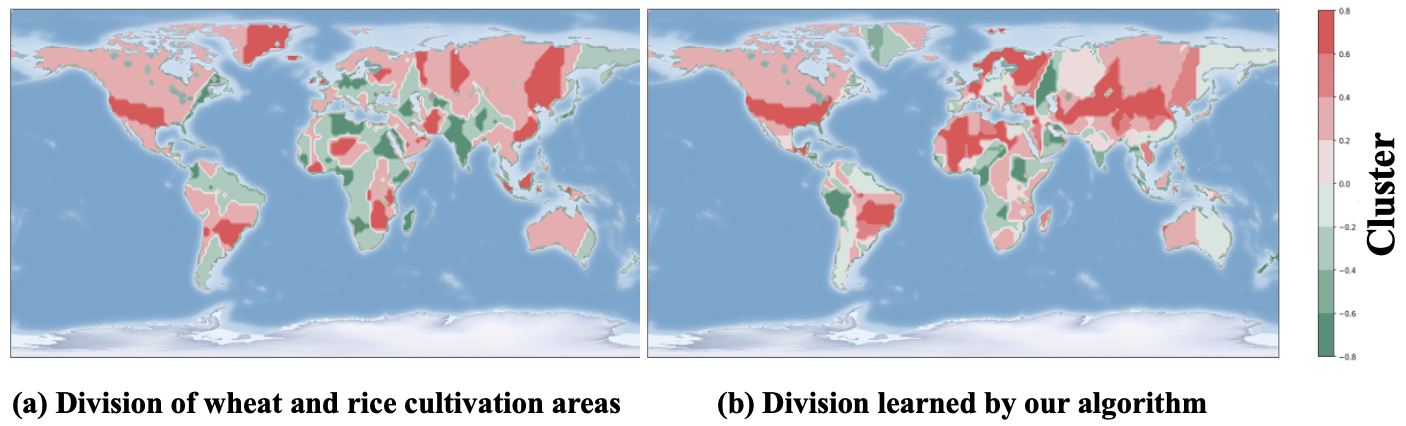}
    \vspace{-0.1in}
	\caption{Results on the crop yield data. Each region is colored according to its main crop type, and the shade represents the proportion of the main crop type after smoothing via $k$-nearest neighbors ($k = 3$). Figures are from \cite{liu2023}.}
	\label{fig:agriculture}
\end{figure*}

Intuitively, Equation~(\ref{equ:usable-predictive-heterogeneity-1}) quantifies the \emph{maximal usable information gain} achievable by dividing the whole dataset $\P(X,Y)$ into several environments $\P(X,Y|E)$.
Consider a collected dataset at hand: if segmenting the dataset significantly enhances the predictive power for the target variable $Y$, this suggests the presence of diverse predictive mechanisms $X\rightarrow Y$ across the partitions.
Based on this notion, the finite sample bounds \cite[Theorem 3]{liu2023} as well as the tractable optimization algorithm \cite[Equation 15]{liu2023} are derived.
Predictive heterogeneity provides valuable insights into the underlying structure of collected data.
We demonstrate its practical utility through applications in predictive tasks within healthcare and agricultural research.

\paragraph{Application 1: Healthcare}
Using a COVID-19 dataset of Brazilian patients~\cite{baqui2020ethnic}, we investigate the task of predicting mortality based on a diverse set of risk factors, including comorbidities, symptoms, and demographic characteristics.
Figure~\ref{fig:covid} presents the results derived from the predictive heterogeneity measure.
Figure~\ref{fig:covid}(a) reveals a clear distinction in the age distributions across the identified subgroups. Specifically, Group 0 is predominantly composed of individuals over the age of 70, Group 1 centers around individuals in their 60s, and Group 2 includes a broader range of middle-aged individuals spanning multiple age brackets.
More importantly, Figure~\ref{fig:covid}(b) shows the average values of various risk factors, highlighting substantial differences among the subgroups—differences that point to distinct underlying causes of mortality.
Group 0 exhibits a significantly higher prevalence of chronic conditions such as renal, neurological, liver diseases, and immunosuppression compared to the other groups. In contrast, Group 1 displays relatively low levels of underlying health issues.
Interestingly, Group 2 does not present any major underlying diseases but shows elevated levels of gastrointestinal symptoms such as diarrhea and vomiting.
These findings underscore that the identified subgroups are characterized by distinct risk profiles associated with COVID-19 mortality. Such insights can support tailored and effective clinical interventions by aligning treatment strategies with subgroup-specific risk factors.

Beyond healthcare, predictive heterogeneity also plays a critical role in agricultural modeling, as illustrated below.
\paragraph{Application 2: Agriculture.}
The task is to predict crop yield at each location using summary statistics of local weather conditions and location-specific covariates (i.e., longitude and latitude)\cite{lobell2008prioritizing}.
Figure~\ref{fig:agriculture}(a) shows the actual geographic distribution of wheat and rice planting areas, while Figure~\ref{fig:agriculture}(b) shows the two sub-populations identified by the predictive heterogeneity method~\cite{liu2023}.
By comparing Figures~\ref{fig:agriculture}(a) and (b), we observe a strong alignment between the learned sub-populations and the true division of crop types. Notably, the crop type information is not available during training and is not used as an input feature in the prediction task.
Given that the mechanisms underlying crop yield predictions are inherently tied to crop type, the close correspondence between the learned sub-populations and actual crop divisions provides compelling evidence that the method effectively captures distinct predictive mechanisms.

In these practical applications, predictive heterogeneity enables the identification of sub-populations governed by distinct prediction mechanisms, thereby enhancing our comprehension of the collected data. 
Additionally, the discovery of varied prediction mechanisms suggests that relying on a singular model may no longer be sufficient (also, invariance assumptions should not be made). 
A more effective strategy involves either augmenting the dataset with additional features or employing multiple models to accommodate the data's complexity. 

In addition to predictive heterogeneity, research in economics by Fan et al. \cite{fan2022we} has demonstrated the heterogeneous patterns in intervention effects, revealing variations in elasticities and optimal pricing across municipalities. 
Furthermore, these impacts vary among different products. This underscores the prevalence of data heterogeneity across a broad spectrum of real-world applications.
This underscores the importance of a thorough understanding of the collected data in guiding subsequent model design and training phases.

\section{Model Training}
\label{sec:training}

The second stage of the ML pipeline is model training, encompassing both the algorithmic design and the training processes subsequent to data collection. 
To maximize data utilization and mitigate biases during model training, a careful approach to handling data is needed. 
Consequently, this section is structured around a pivotal question:
\begin{center}
\textbf{Q2: How to utilize data effectively and judiciously?}
\end{center}
In response, we will present a selection of studies that illustrate the significance of incorporating data heterogeneity into the model training process as a means to address this question. 
These works highlight the benefits of nuanced data handling, providing insights into improving model generalization performance.

In traditional machine learning approaches, models are typically optimized through Empirical Risk Minimization (ERM), where data from different sources is pooled together, and the average risk across the dataset is computed.
However, when using ERM, the learned models are prone to neglect minority groups, thereby introducing issues related to fairness, bias, and reliability.
This underscores the necessity for establishing a new framework that ensures both effective and judicious use of data during the model training phase.

Current research primarily tackles these challenges from a model-centric or algorithm-centric perspective, leading to diverse algorithmic branches aimed at enhancing reliability and generalization. 
This includes approaches such as Distributionally Robust Optimization \cite{blanchet2019data, blanchet2019robust, sagawa2019distributionally, duchi2021learning, liu2025dropythonlibrary}, Invariant Learning \cite{arjovsky2019invariant, ahuja2020invariant, ahuja2021invariance, chen2022pareto}, and Domain Generalization \cite{shen2021towards,zhou2022domain,wang2022generalizing}, among others.
Despite the wealth of methodologies introduced, as previously discussed in Section \ref{sec:pre}, their efficacy in practical applications continues to be questioned.

\begin{figure}
	\vspace{-0.2in}
	\includegraphics[width=0.8\linewidth]{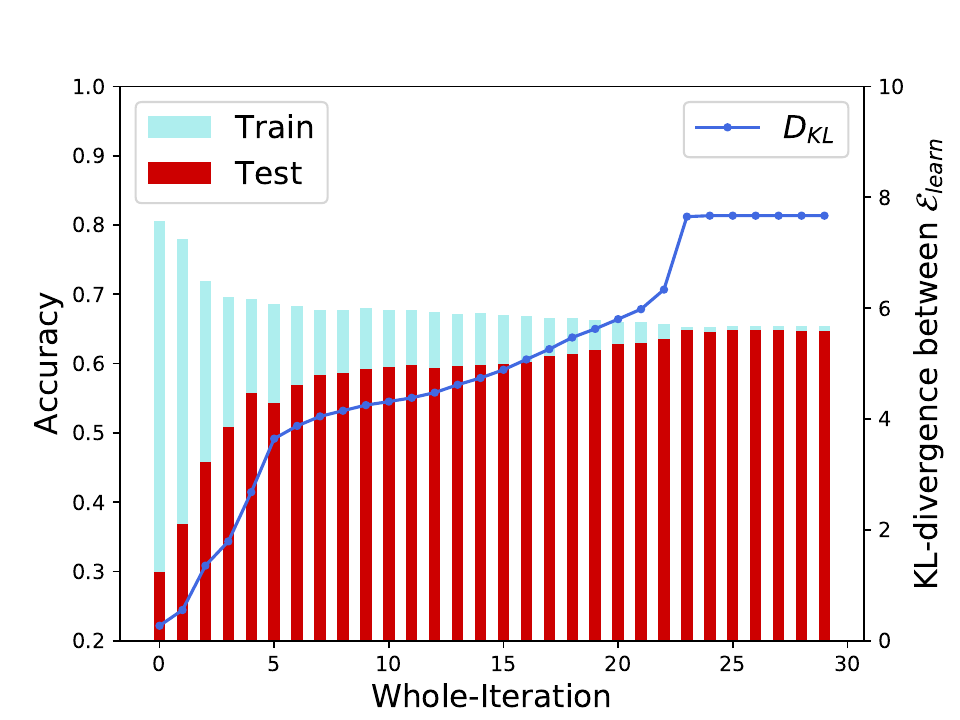}	
    \vspace{-0.1in}
	\caption{Case study by Liu et al. \cite{liu2021kernelized} that demonstrates better learned sub-populations lead to better generalization performances. Figure from \cite{liu2021kernelized}.}
	\label{fig:kerhrm}
	\vspace{-0.2in}
\end{figure}

In this section, under the framework of heterogeneity-aware machine learning, we concentrate on an emerging and promising research direction. 
This approach simultaneously investigates the underlying structures and information within data and leverages these insights during model training.
The central idea emphasizes actively uncovering the inherent heterogeneity within data, rather than relying on unfounded assumptions.
We refer the readers interested in other branches of methods to this survey \cite{shen2021towards}.

\subsection{Explicit Modeling}
We first introduce algorithms that explicitly model data heterogeneity throughout the training process and harness this identified heterogeneity to enhance model training.
The concept of explicitly addressing data heterogeneity during model training was first introduced through Heterogeneous Risk Minimization (HRM) \cite{liu2021heterogeneous}.
HRM incorporates two synergistic modules: a backend module that utilizes the learned sub-populations to distinguish between stable features $S$ and unstable features $V$ within the input covariates $X$, and a frontend module that employs a Gaussian mixture model to delineate sub-populations with different $\P(Y|V)$ distributions.
These modules are jointly optimized, facilitating mutual benefits. 
Specifically, as the frontend module identifies more accurately sub-populations, the backend module can pinpoint unstable features $V$ more effectively. 
Consequently, this improved identification of $V$ enables the frontend module to refine its delineation of sub-populations, thus creating a cycle of enhancement.
Building on this, Liu et al. \cite{liu2021kernelized} extend the approach by integrating the Neural Tangent Kernel (NTK) to accommodate more complex data types, such as images.
As illustrated in Figure \ref{fig:kerhrm}, along with the joint optimization, the model's generalization performance (indicated by the red bars) improves consistently when the learned sub-populations exhibit greater heterogeneity (as shown by the blue curve).
Also, Xu et al. \cite{xu2022distributed} considers the data heterogeneity problem in a distributed learning setting, and they allow for distinctive function maps for data scattered at different locations. 

\begin{definition}[$\alpha_0$-distributional stability \cite{liu2024enhance}]
\label{def:distributionalinvariance}
Given distribution $\mathbb P(Z)$, for $\alpha_0\in(0,1/2)$ as a lower bound  on the proportion $\alpha$, the set of sub-populations of distribution $\mathbb P$ is $\mathcal{P}_{\alpha_0}(\mathbb P)\coloneq \{\mathbb Q_0: \mathbb P=\alpha \mathbb Q_0+(1-\alpha)\mathbb Q_1,\text{ for some } \alpha\in[\alpha_0,1) \text{ and distribution }\mathbb Q_1 \text{ on }\mathcal{Z} \}$.
The $\alpha_0$-distributional stability of the prediction mechanism $Y|X$ is defined as:
\begin{equation}
\label{equ:def}
\text{DS}_{\alpha_0}(Y|X;\mathbb P)\coloneq	\sup_{\mathbb Q\in\mathcal{P}_{\alpha_0}(\mathbb P)}\rho_{\text{KL}}(\mathbb Q(Y|X), \mathbb P(Y|X)),
\end{equation}
where $\rho_{\text{KL}}(\cdot,\cdot)$ denotes the KL-divergence between two distributions.
\end{definition}
Recently, \citet{liu2024enhance} have formally introduced the concept of \(\alpha_0\)-distributional stability with respect to shifts in \(Y|X\), as outlined in Definition \ref{def:distributionalinvariance}. This measure quantifies the maximum variation in \(Y|X\) across all sub-populations exceeding a size of \(\alpha_0\). Furthermore, \(\text{DS}_{\alpha_0}(Y|X;\mathbb P)\) is employed as a penalty mechanism to ensure model robustness within a defined range of sub-population shifts.
Furthermore, Huang et al. \cite{huang2022environment} utilize multi-head neural networks for the efficient inference of sub-populations, particularly in handling complex data scenarios.
Moreover, this area of research has spurred applications in graph data analysis and recommendation systems.
\paragraph{Application 3: Graph data.}
\citet{wu2021handling,gui2024joint,chen2024does} have adapted the HRM framework to graph data by implementing several modifications: (1) substituting the original stable/unstable features with stable/unstable sub-graphs, (2) replacing the feature selection module with various edge/node selection or generation techniques, and (3) employing alternative forms of penalties in place of the traditional risk penalty.
These approaches have shown impressive performance across diverse graph datasets with distribution shifts, underscoring the practical benefits of explicitly modeling data heterogeneity during the training process.

\paragraph{Application 4: Recommendation systems.} 
The HRM concept has been successfully applied to recommendation systems as well.
Wang et al. \cite{wang2022invariant} adapt the HRM framework to parse invariant and variant preferences from biased observational user behaviors. 
Their algorithm demonstrates a significant capability for debiasing. 
Similarly, Du et al. \cite{10.1145/3503161.3548405} focus on distinguishing sub-populations through unique user-item interactions present within the dataset, incorporating this detailed insight into the training process. 
This strategy not only minimizes spurious correlations but also provides a more accurate reflection of genuine user preferences.

\subsection{Implicit Modeling}
Contrary to the algorithms discussed earlier, which explicitly model heterogeneity by identifying discrete sub-populations, there are methods adopting an implicit strategy for addressing data heterogeneity during the model training stage. 
The term "implicit" refers to the approach of not splitting the dataset into sub-populations; instead, these methods direct the models to focus more on sub-populations that exhibit lower performance. 
Originating from the principles of Distributionally Robust Optimization (DRO), this approach uses a more data-driven way to mitigate the limitations identified in previous discussions (as outlined in Section \ref{sec:pre}).
Blanchet et al. \cite{blanchet2019data} and Liu et al. \cite{liu2021stable,liu2022distributionallya} employ metric learning techniques to define the geometric properties of data, enhancing the Distributionally Robust Optimization (DRO) approach by concentrating on a more practical worst-case distribution. 
Building upon this, Liu et al. \cite{liudistributionally} apply a $k$-nearest neighbor graph to approximate the data manifold. 
They then construct the ambiguity set using the geometric Wasserstein distance, which prompts the model to focus on the underperforming sub-populations that are smooth along the manifold.
Building on this, subsequent research \cite{liu2023geometry} introduces geometric calibration terms to more effectively tackle the issue of noise within DRO, and \citet{zheng2024topology} achieve superior generalization performance on graph data.
In life science domain, \citet{fan2024stable} apply sample reweighting in survival analysis to identify robust biomarkers across diverse patient populations.

These studies and applications offer insightful perspectives on applying the idea of exploring heterogeneity during model training, showcasing the potential and practical promise of this innovative paradigm.

\begin{figure*}[htbp]
\vspace{-0.1in}
	\includegraphics[width=0.8\textwidth]{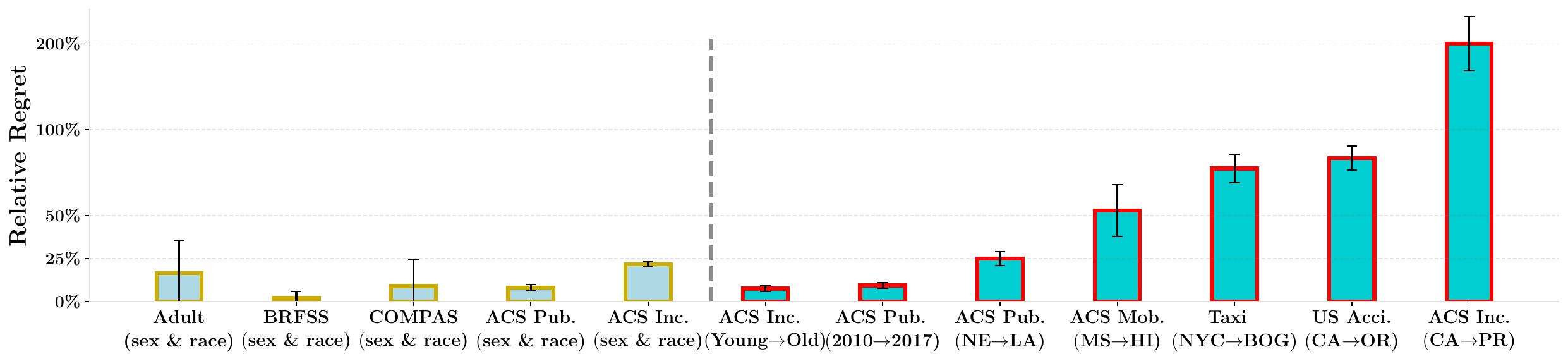}
    \vspace{-0.1in}
	\caption{Relative regret in typical benchmarks \cite{Dua:2019, gardner2022subgroup}  (left 5 bars) and seven settings designed in WhyShift benchmark \cite{liu2024need} (right 7 bars). Figure from \cite{liu2024need}.}
	\label{fig:regret}
    \vspace{-0.2in}
\end{figure*}

\section{Model Evaluation}
\label{sec:evaluation}

The third stage in the ML pipeline is model evaluation, a critical phase where ML engineers need to assess the performance of their trained models prior to deployment. 
This step ensures that the models meet the expected standards of accuracy, fairness, and reliability, aligning with the objectives set out at the project's inception. 
Therefore, it is of paramount importance to use right ways of evaluation.
This involves a comprehensive analysis using various metrics and validation techniques to identify any potential issues or areas for improvement, thereby safeguarding against unintended consequences when the models are finally put into use.
In this section, we focus on the following question:
\begin{center}
	\bf Q3: How to ``actively'' evaluate models with right data?
\end{center}
We aim to offer insights into how a better understanding of data heterogeneity can enhance model evaluation from both perspectives: selecting the appropriate data and employing the ``active'' metrics. 
This approach underscores the importance of nuanced evaluation strategies in achieving comprehensive and accurate model assessments.

\subsection{Right Evaluation Data}
The most intuitive approach for evaluation involves utilizing data that is drawn independently and identically distributed ($i.i.d.$) from the training distribution, typically through cross-validation methods.
Widely used in traditional machine learning, it enables the assessment of a model's performance on the training distribution and helps prevent overfitting.
However, in real-world applications, the target data distribution frequently diverges from the training distribution, and ML models often encounter post-deployment performance degradation.
For instance, in the context of predicting house prices, a model that performs well on historical data often experiences a significant decline in performance when applied to new, unseen data \cite{shen2020stable}; for insurance prediction, the model performance varies significantly across different US states and demographic groups \cite{ding2021retiring}.
Therefore, prior to model evaluation, it is crucial to ensure that the data aligns with the objectives of the evaluation.

Beyond the scope of $i.i.d.$ scenarios, to accurately evaluate a model's generalization capabilities in the face of distribution shifts, numerous benchmarks across a variety of domains have been introduced, e.g., PACS \cite{li2017deeper}, NICO++ \cite{zhang2022nico++} for image data, Retiring Adult \cite{ding2021retiring}, Bank Account Fraud dataset \cite{jesus2022turning} for tabular data, and WILDS \cite{koh2021wilds} for multiple data types etc. 
Nonetheless, these benchmarks often overlook the specific patterns of distribution shifts, with researchers not sufficiently understanding the data heterogeneity.
Given that algorithms tailored for certain type of distribution shifts ought to be validated on data with corresponding shift patterns, simply relying on these benchmarks may result in evaluations that do not accurately reflect the algorithm's effectiveness.

To explore this thoroughly, in the context of tabular data, Liu et al. \cite{liu2024need} introduce relative regret (as defined in Equation \eqref{eqn:regret}) to examine the patterns of distribution shifts between \(\P\) and \(\Q\).
\begin{equation*}
  \label{eqn:regret}
  \frac{\mathbb{E}_\Q[\ell(Y, f_\P(X))]}{\min_{f\in \Fscr} \mathbb{E}_\Q[\ell(Y, f(X))]} - 1,
  ~~~\mbox{where}~~~
  f_\P \in \arg\min_{f\in \Fscr} \mathbb{E}_\P [\ell(Y, f(X))],
\end{equation*}
where $\ell(\cdot,\cdot)$ is the 0-1 loss.
For widely-used benchmarks \cite{Dua:2019, gardner2022subgroup}, the relative regret is small (left 5 bars in Figure \ref{fig:regret}), suggesting the $Y|X$ distribution is largely transferable across those groups.
The 7 selected settings in WhyShift benchmark \cite{liu2024need} consist of prediction tasks from different fields, such as income prediction, insurance prediction, and accident prediction.
The relative regrets of the 7 settings in WhyShift vary a lot (right bars in Figure \ref{fig:regret}), indicating a wide range of $Y|X$-shifts. 
Figure \ref{fig:regret} shows that the commonly-used tabular datasets primarily contain covariate shifts (\(X\)-shifts) while neglecting shifts in \(Y|X\).
Additionally, Liu et al. \cite{liu2024need} reveal through comprehensive empirical studies that the ``accuracy-on-the-line'' phenomenon \cite{miller2021accuracy}, observed across numerous image datasets, fails to persist in the presence of significant \(Y|X\)-shifts.
Similarly, Yang et al. \cite{yang2023change} decompose the sub-population shifts into four types, i.e. spurious correlations, attribute imbalance, class imbalance, and attribute generalization, and provide fine-grained benchmarking results.
And the attribute generalization problem is found much more challenging among sub-population shifts.

These recent benchmarks highlight the role of a deeper comprehension of data heterogeneity in selecting the appropriate data for evaluation. 
For instance, datasets primarily characterized by covariate shifts are not advisable for evaluating algorithms intended to address \(Y|X\)-shifts.

\subsection{Active Evaluation Algorithm}
Beyond evaluation using ``static'' datasets, a more promising approach is to assess models with actively generated data—an increasingly important strategy in the era of large language models, where data contamination poses a significant challenge to reliable evaluation.

To this end, \citet{blanchet2024stability} propose a principled approach to evaluating model robustness under distribution shifts.
Instead of relying on static test sets, this framework focuses on quantifying the stability of a trained model by measuring the minimal perturbation to the data distribution required to cause a prescribed degradation in performance. The method models perturbations in a unified way through optimal transport (OT) with moment constraints over a joint sample-density space, allowing it to capture both data corruptions (perturbations to the support) and sub-population shifts (changes in probability mass). The resulting optimization problem is equipped with strong duality guarantees and leads to tractable formulations across different classes of loss functions. 
This framework offers a general and theoretically grounded tool for understanding how machine learning models behave under realistic distributional perturbations.

In addition to evaluating model stability, several studies~\citep{eyubogludomino, johnson2023does, liu2024need, ghosh2024ladder} focus on identifying regions where a model underperforms---a task commonly referred to as \emph{error slice discovery}. 
An \emph{error slice} refers to a subset of data samples that exhibit poor model performance and share common characteristics. 
For example, \citet{eyubogludomino} propose a method that combines cross-modal embeddings with an error-aware mixture model to uncover and describe coherent error slices. 
Similarly, \citet{liu2024need} use regression tree models to identify high-risk regions characterized by significant shifts in the conditional distribution $\mathbb{P}(Y \mid X)$, while \citet{yu2025error} introduce compactness regularization to further constrain the structure of discovered slices. 
In a related line of work, \citet{thams2022evaluating} integrate domain knowledge into a causal graph and construct a parametric robustness set of distributions. Leveraging this set, they develop a second-order approximation algorithm to detect significant and plausible shifts that could adversely affect model performance. 
Overall, identifying these risky regions or patterns supports the development of interpretable rules that define covariate subpopulations, thereby enhancing our understanding of where and why model failures are most likely to occur.

\paragraph{Application 5: Large Language Models.}
Due to the growing concerns around data contamination, the \emph{reliable} evaluation of large language models (LLMs) is becoming increasingly challenging. These challenges stem from issues related to both the evaluation datasets and the evaluation methodologies. 
As more benchmarks are introduced, more effort has been directed toward improving the quality and diversity of test data. 
For instance, \citet{xiatop} extend existing benchmarks to cover various targeted domains, enabling a more comprehensive assessment of LLMs’ coding capabilities. Similarly, \citet{whitelivebench, chen2025memorize} propose the use of evolving questions to better evaluate the generalization and robustness of LLMs.

\section{Model Deployment}
\label{sec:deployment}
The fourth stage of the machine learning pipeline is model deployment, where the trained model is integrated into real-world applications. At this point, performance degradation may become apparent as the model encounters data that differ from its training distribution.
Although the model is now operational, a key challenge remains: analyzing failure cases and efficiently updating the model. 
This section examines how understanding data heterogeneity can shed light on the central question:
\begin{center}
\bf Q4: How can we diagnose model's performance degradation for efficient model improvement?
\end{center}
This involves two critical steps: (1) identifying the root causes of failure, and (2) designing targeted interventions—both from data-centric and algorithmic perspectives.

When encountering performance degradation in the target distribution, attributing the drop to specific types of distribution shifts is crucial, as each type necessitates a unique solution. 
For instance, shifts in \(X\) can result from temporal changes, changes in population demographics, among others. 
To deal with \(X\)-shifts, methods like importance weighting \cite{asmussen2007stochastic, shen2020stable, duchi2023distributionally} and domain adaptation \cite{shimodaira2000improving, ganin2015unsupervised} may be beneficial. 
On the other hand, shifts in \(Y|X\) could stem from measurement errors or unobserved confounders. 
In such cases, additional data collection and labeling become essential for addressing these shifts effectively.
To this end, Cai et al. \cite{cai2023diagnosing} categorize distribution shifts into two types: shifts in the marginal distribution of the covariates ($X$-shifts), and shifts in the conditional relationship between the outcome variable and covariates ($Y|X$-shifts).
They introduce a hypothetical shared distribution on $X$, comprising values common in both the training and target distributions. 
This shared distribution facilitates the comparison of $Y|X$, thereby allowing for the decomposition of $X$-shifts versus $Y|X$-shifts.
In a subsequent study, \citet{singh2024hierarchical} delve deeper into the problem by incorporating Shapley values to enable a hierarchical decomposition of distribution shifts. 
To address the computational cost associated with Shapley values, \citet{liugoing} introduce the \textit{feature attribution score}, which is analogous to the average treatment effect, defined as:
\begin{equation}
\label{equ:attribution}
    \text{Attr.}(S) \coloneq \mathbb{E}\left[R_{\mathbb{Q}}(X_{-S}) - R_{\mathbb{P}}(X_{-S})\right],
\end{equation}
where \( R_{\mathbb{P}}(X_{-S}) \coloneq \mathbb{E}_{\mathbb{P}}\left[\ell(f(X), Y) \mid X_{-S}\right] \), and similarly for \( R_{\mathbb{Q}}(X_{-S}) \). This score captures the \textit{performance gap} between distributions \( \mathbb{Q} \) and \( \mathbb{P} \), while conditioning on the marginal distribution of all features except those in the subset \( S \), denoted by \( X_{-S} \).
From the perspective of distribution shift, \( \text{Attr.}(S) \) quantifies the performance degradation caused by changes in the conditional distribution \( (Y, S) \mid X_{\setminus \{S\}} \) between \( \mathbb{P} \) and \( \mathbb{Q} \).

\paragraph{Application 6: Health Care \& Finance}
Post-deployment performance diagnosis is essential in high-stakes domains such as health care, where timely detection and mitigation of model failures can directly impact patient outcomes and potentially save lives. 
In the context of ICU mortality risk prediction, \citet{liugoing} show that a detailed understanding of the features experiencing the most significant distribution shifts enables a simple group-balancing strategy to outperform commonly used fine-tuning methods on limited test data, resulting in improved generalization performance.

In the financial domain, \citet{cai2023diagnosing, liu2024need} demonstrate that collecting relevant features guided by shift attribution can substantially reduce generalization gaps. 
Similarly, \citet{zhou2022measuring} incorporate land-use data to address data heterogeneity in housing vitality analysis, highlighting the importance of domain-specific information in improving model reliability.
\citet{xu2025heterogeneous} focus on the competition among model providers in heterogeneous data markets and analyze the resulting equilibrium, offering insights into policy design that fosters fair and diverse model deployment.

Notably, the model performance diagnosis offers valuable insights for targeted data collection efforts, thereby completing the cycle of the machine learning pipeline.

\section{Conclusion}

Throughout this paper, we have provided a comprehensive overview of heterogeneity-aware machine learning, spanning the four critical stages of the machine learning pipeline. 
To conclude, we reflect on a central question: What lies ahead for the future of heterogeneity-aware machine learning?

\paragraph{Theoretical Foundations.}
As emphasized by \citet{cai2023diagnosing, liu2024need}, there remains a significant gap in establishing a unified modeling framework for data heterogeneity. 
Although several definitions have been proposed~\citep{liu2024need, liu2024enhance}, a common foundation for studying and discussing data heterogeneity is still lacking—despite it being a fundamental and intrinsic property of large-scale data. 
Our case studies further underscore the deep connection between data heterogeneity and critical issues such as model fairness and bias. 
In many ways, data heterogeneity can be viewed as a dual problem to fairness, bias, and generalization, prompting a shift from a model-centric to a more data-centric paradigm. 
Establishing solid theoretical foundations for this area represents a promising and necessary direction for future research.

\paragraph{Empirical Power.}
This paper has highlighted the promising benefits of heterogeneity-aware machine learning across a range of critical domains, including health care, agriculture, finance. These case studies illustrate how accounting for data heterogeneity can lead to more robust, fair, and generalizable models. However, the empirical demonstrations thus far have primarily focused on relatively small-scale or domain-specific applications. 
To fully unlock the potential of the heterogeneity-aware machine learning paradigm, future efforts must scale these methodologies to broader and more complex settings. In particular, large-scale applications such as large language models (LLMs), foundation models, and multimodal systems present both new challenges and opportunities. These systems are trained on vast, diverse datasets that inherently contain significant heterogeneity across domains, modalities, and user contexts. Incorporating heterogeneity-aware techniques into their training, evaluation, and deployment pipelines could yield substantial gains in performance, fairness, and adaptability.
Realizing this vision will require not only methodological advancements but also the development of scalable toolkits, benchmark datasets, and evaluation protocols that explicitly consider heterogeneity. The empirical power of this paradigm will be most fully demonstrated when these principles are integrated into real-world systems.

We encourage more researchers to engage with this important line of inquiry. 
We invite the community to further explore and advance the frontiers of heterogeneity-aware machine learning and exploring its vast potential to transform how we understand, develop, and deploy reliable machine learning systems across a wide range of disciplines and real-world applications.



\newpage

\begin{acks}
This work was supported by NSFC (No. 62425206), Tsinghua-Toyota Joint Research Fund, NSFC (No. 62141607), and Beijing Municipal Science and Technology Project (No. Z241100004224009).
We would like to express our sincere gratitude to Bo Li for the
initial inspiration behind the term ``data heterogeneity''. Our deep appreciation is extended to Hongseok Namkoong, Jose Blanche, Jiajin Li and Mihaela van der Schaar for their invaluable guidance on model evaluation. Furthermore, we thank Tiffany Tianhui Cai, Zheyuan Hu, Zheyan Shen, Nabeel Seedat, Tianyu Wang, Zimu Wang, Jiayun Wu, and Han Yu for their contributions. 
\end{acks}

\bibliographystyle{ACM-Reference-Format}
\bibliography{sample-base}


\end{document}